%% file: main.tex
\input{preamble.tex}
\mlsystitlerunning{State-Aware Activation Recomputation}
\begin{document}
\twocolumn[
\mlsystitle{State-Aware Activation Recomputation with Replay Contracts and Selective Preservation}
\begin{mlsysauthorlist}
\mlsysauthor{Annan Zhang}{pku}
\mlsysauthor{Shengjie Song}{pku}
\mlsysauthor{Wenshuai Yao}{pku}
\mlsysauthor{Lihao Zhang}{pku}
\mlsysauthor{Tianmiao}{buaa}
\end{mlsysauthorlist}
\mlsysaffiliation{pku}{Peking University}
\mlsysaffiliation{buaa}{Beihang University}
\mlsyscorrespondingauthor{Annan Zhang}{\mbox{annan@pku.edu.cn}}
\vskip 0.3in
\begin{abstract}
\input{abstract}

\end{abstract}
]
\printAffiliationsAndNotice{}
\input{body}

\clearpage
\bibliography{references}
\bibliographystyle{mlsys2025}
\clearpage
\appendix
\section{Complete Cost-Calibration Grid}
\label{app:grid}
Table~\ref{tab:cost-calibration-grid} lists every shape and method of the prespecified H100 cost calibration summarized in the main text.
\input{cost_calibration_grid}
\end{document}

%% file: preamble.tex
\documentclass{article}
\usepackage{microtype}
\usepackage{xcolor}
\usepackage{hyperref}
\usepackage{mlsys2025}
\renewcommand{\icmlruler}[1]{}
\usepackage{etoolbox}
\makeatletter
\patchcmd{\mlsysauthor}{\ifdefined\isaccepted}{\iftrue}{}{%
  \PackageError{authordetails}{Could not show author names}{Check the MLSys style.}}
\patchcmd{\mlsysaffiliation}{\ifdefined\isaccepted}{\iftrue}{}{%
  \PackageError{authordetails}{Could not show affiliations}{Check the MLSys style.}}
\patchcmd{\mlsyscorrespondingauthor}{\ifdefined\isaccepted}{\iftrue}{}{%
  \PackageError{authordetails}{Could not show the corresponding author}{Check the MLSys style.}}
\makeatother
\usepackage{xurl}
\usepackage[T1]{fontenc}
\usepackage[utf8]{inputenc}
\usepackage{amsmath,amssymb,mathtools}
\usepackage{booktabs,tabularx,array,multirow}
\usepackage{graphicx,xcolor}
\usepackage{placeins}
\newcolumntype{Y}{>{\raggedright\arraybackslash}X}

%% file: abstract.tex
Activation recomputation saves accelerator memory by replaying forward segments during the backward pass, but replay is only correct if it sees the state the original forward pass saw. Mutable buffers, aliased bytes and random-generator state can change in between, so re-executing unchanged code can differentiate a different computation. We make this state explicit. A versioned replay contract specifies what replay must restore, and a joint planner chooses, per segment, between activation retention, whole-buffer snapshots and selective-region preservation, charging each option its measured memory and recovery cost. Independent derivative and byte-level checks validate the contract, and a Torch/JAX cohort retains all 5,900 execution records, including matches, rejections and divergences. On frozen sparse-write H100 workloads, each measured over fifty fresh processes, joint planning reduces paired step time by 18.16\% against a correctly costed eager/whole planner at a 44 GiB budget (approximate 97.5\% ratio interval $[.8172,.8195]$). At 32 GiB, where the eager/whole planner has no feasible static policy, joint planning is 4.98\% faster than fixed regions at 29.94 rather than 26.20 GiB. A complete CIFAR-100 dense-queue application runs seven methods for 100 encoder and 25 probe epochs at six training seeds, with byte-identical test scores within each seed. There, joint planning uses 22.41\% more allocated memory than whole snapshots, with a process-time geometric-mean ratio of 1.0080 (seed range $[.9845,1.0191]$). Together, the results map where selective preservation pays off: sparse writes with large activations benefit, while in the dense-queue workload whole snapshots use less allocated memory.

%% file: body.tex
\section{Introduction}
Activation recomputation lets training fit in limited accelerator memory by discarding activations and replaying forward segments during the backward pass. Replay is only correct if it reproduces the execution whose gradient is required. Mutable buffers can change before replay, aliases can expose the same bytes through different indices, and random draws can be absent from a tensor dependency graph. Re-executing unchanged code can therefore differentiate a different computation.

Correctness and memory planning are therefore coupled. Whole-state snapshots make replay valid but are costly. Selective preservation reduces that cost when writes are sparse, and a planner that ignores historical-state memory and recovery can exceed its own budget. We specify a versioned replay contract, implement declared-state preservation, and plan activation and state retention jointly from measured profile cards. Finite-program derivatives and byte traces test the contract, and a Torch/JAX study retains all 5,900 matches, rejections and divergences.

The evaluation identifies the regime in which joint planning pays off. In the original 24-shape calibration and a completed 150-process, three-workload cohort, eager retention is already both faster and smaller. In separate fifty-process sparse-write cohorts, joint planning is 18.16\% faster than a correctly costed eager/whole planner at 44 GiB. At 32 GiB, it selects a policy that the eager/whole planner cannot fit, trading speed against memory relative to fixed regions. Budgets are logical rather than enforced, so allocation above a budget is not a device out-of-memory event.

\input{teaser_figure}
Figure~\ref{fig:teaser} summarizes the headline measurement. Our contributions are:
\begin{itemize}\setlength{\itemsep}{1pt}\setlength{\parskip}{0pt}
\item A versioned replay contract that identifies values by their producing write and requires replay to reproduce both storage bytes and random-generator state (Section~\ref{sec:contract}).
\item Declared-state preservation engines, including selective region journals that cover the union of current and later destroying writes (Section~\ref{sec:contract}).
\item A joint planner that charges activation retention, temporary workspace and preservation/recovery cost in one static model, with untruncated Pareto frontiers (Section~\ref{sec:planning}).
\item Executed evidence at several scales: 1,024-program derivative and byte suites, a 5,900-record Torch/JAX cohort, and fifty-process H100 sparse-write cohorts showing an 18.16\% step-time reduction at 44 GiB and a 4.98\% speedup over fixed regions at 32 GiB.
\item A six-seed CIFAR-100 dense-queue application in which all seven methods produce byte-identical test scores within each seed, which locates the boundary of the sparse-write benefit.
\end{itemize}

In the application, all seven methods train for 100 encoder and 25 probe epochs at six independent training seeds. Full-process observation at five seeds shows that whole snapshots use less allocated memory at every seed, while the encoder-time ordering reverses at one seed. The sparse-write benefit and the dense-queue resource tradeoff together define where each preservation strategy should be chosen.

\section{Related Work}
Automatic differentiation composes derivatives of executed operations \citep{baydin}. Enzyme applies differentiation through compiler intermediate representations \citep{enzyme}. We expose temporal state dependencies and test declared-state replay; extending this to a general compiler transformation is complementary work.

Checkpointing trades retained activations for recomputation. Sublinear-memory training \citep{chen}, dynamic programming for backpropagation through time \citep{gruslys}, and schedule optimization in Checkmate \citep{checkmate} motivate the planning component. Our planner explicitly charges for the historical mutable state that makes a segment's replay valid.

Coop considers allocation, fragmentation and in-place computation \citep{coop}; the static profile model here does not establish allocator-level optimality. PyTorch supplies a broader execution and differentiation environment \citep{paszke2019pytorch}. The framework checker isolates the consequences of hidden state and admission boundaries rather than treating unsupported stateful programs as violations of a pure-function checkpoint API.

Reversible residual networks recover activations through architectural structure \citep{revnet}; reversible learning differentiates training procedures \citep{reversible}; neural ordinary differential equations use a continuous-state formulation \citep{neuralode}. Each requires a particular recovery contract. ZeRO-Offload adds a placement and communication tradeoff \citep{offload}, a natural extension of the static policy space.

\section{The Replay Contract}
\label{sec:contract}
Let $B$ denote physical storage bytes and $R$ random-generator state. A logical value is identified by its producing write, not solely by its address. Every operation reads the versions current before its writes; overlapping source/destination sets therefore cannot be resolved by consulting the final mutable buffer.

The finite numerical implementation creates a new node for each write, retaining the version identifiers of its operands, coefficients, sampled noise and nonlinearity. For $v_k=f_k(v_{p(k)})$, reverse differentiation accumulates
\begin{equation}
 \overline v_{p(k)}\mathrel{+}=\overline v_k f'_k(v_{p(k)}).
\end{equation}
An independently organized forward Jacobian supplies the derivative reference. Reusing recorded noise is distinct from drawing a new sample during recomputation.

Byte-level replay separately requires
\begin{equation}
 \mathcal P(B_0,R_0)=(B_T,R_T).
\end{equation}
Both components matter: equal final bytes with a differently advanced generator can change a later execution. Typed aliases share physical storage, so numerical equality of one view is insufficient. The byte suite compares the full storage and generator state; it does not claim differentiability of arbitrary integer bit mutations.

\subsection{Selective state preservation}
Whole snapshots preserve the declared buffer before replay. Region journals retain overwritten values and undo/redo writes in their recorded order. Read-set preservation stores historical operands explicitly. The finite engines and their recovery boundaries are listed in Table~\ref{tab:engine-contracts}.
\input{preservation_engines}

The H100 cost runtime executes real state reads, writes, views, forward/backward computation and updates. To reconstruct a historical segment, its region baseline conservatively preserves the union of its own and later writes that could destroy that state. It includes row indices, gathered values, scatter work and region expansion. Saving only the current segment's overwritten values would be insufficient when a later segment overwrites another needed region.

The framework transaction restores declared hidden buffers and RNG state around recomputation, then reinstates the state visible outside replay. It does not infer arbitrary aliases or make external side effects transactional. Unsupported integer-write differentiation, raw API rejections, concurrent mutation, asynchronous races and undeclared state remain explicit boundaries.

Figure~\ref{fig:stateful-replay-architecture} summarizes the replay transaction and the state-aware planning options.
\input{stateful_replay_architecture_v8}

\section{Planning Activations and State Together}
\label{sec:planning}
For segment $i$, the measured profile exposes eager retention and replay with either whole or region state preservation. Each option $o$ has retained bytes $r_{i,o}$, temporary bytes $q_{i,o}$ and time $c_{i,o}$. Given persistent bytes $P$ and logical budget $M$, the static policy model minimizes
\begin{equation}
 C=\sum_i c_{i,o_i}
 \quad\text{subject to}\quad
 P+\sum_i r_{i,o_i}+\max_i q_{i,o_i}\le M .
\end{equation}
Replay costs include a repeated forward pass and the measured preservation/recovery work. Model predictions select policies; actual allocator peaks and synchronized step times evaluate them.

The dynamic program retains a Pareto frontier in retained memory, maximum temporary memory and cost. Dominance on only their combined memory would be unsafe because a later segment can change the temporary maximum. Without truncation, the frontier gives the optimum within this static option model. The implementation records any truncation and relinquishes that claim when it occurs. The reported activation-pressure frontiers are untruncated.

Two controls separate sources of benefit. The state-agnostic planner selects eager or whole replay while omitting preservation memory/time from its objective; runtime still preserves state correctly. A stronger state-aware control uses the same correct costs as joint planning but permits only eager and whole snapshots. Comparing the latter with joint planning isolates the value of selective regions; comparing the former exposes the consequence of omitting state from the planning model.

\input{framework_torch_results}

\section{Independent Derivative and Byte Checks}
An initial 1,024-program numerical suite uses dimensions 4/8/16 and 8/12/16 operations, with overlapping and strided writes and recorded randomness. The version graph reproduces scalar outputs exactly; maximum gradient disagreement with the forward Jacobian is $8.88\times10^{-16}$. A control that reads the final storage instead of historical versions is detected on every program.

A separate 1,024-program byte suite checks exact bytes and generator state. Both distinct valid paths---snapshot replay and region undo---match every input. The historical checkpoint label shares the snapshot implementation and is not counted as a third independent engine; all original labeled records remain retained. Replaying from the wrong RNG state is detected on every input. Numerical derivatives and byte-alias behavior are different checks, not interchangeable evidence for differentiation through arbitrary bytes.

NumPy supplies the finite-program arrays and trace analysis \citep{numpy2020}. Small derivative residuals reflect differing floating-point accumulation orders \citep{higham2002}; exact byte equality is required for replay of identical byte operations.
\input{evidence_results}

\input{cost_calibration_results}

\input{activation_pressure_results_v6}
\input{fair44_results_v1}

\input{confirmation_results_v1}
\input{original150_results_v2}

\input{contrastive_four_seed_results_v2}

Figure~\ref{fig:stateful-replay-evidence} places every executed method of the three main cohorts on shared memory and time axes, which makes the regime boundary visible: selective regions pay off under sparse writes and large activations, while in the dense application whole snapshots remain the smaller policy.
\input{stateful_replay_evidence_v8}

\FloatBarrier

\section{Discussion and Limitations}
Selective preservation helps when activations exceed the desired eager budget, writes are sparse, and multiple historical versions make whole snapshots costly. Eager remains competitive when it fits. The state-agnostic planner is an omitted-cost diagnostic, qualifying its 30.08\% memory comparison. The independently confirmed 18.16\% time reduction at 44 GiB isolates selective regions against correctly costed eager/whole policies; fixed regions use less memory at 32 GiB. These static-model results do not exclude other schedules, allocator strategies, offloading or application rewrites.

The fifty-process pressure cohorts and distinct 150-process state-cost cohort address different workload regimes and are not pooled. The latter favors eager retention. The application evaluates six independent training seeds and includes five paired complete-process observations per method. These sample sizes support descriptive contrasts on the stated workload, without a population quality-equivalence claim or application performance confidence interval.

Declared-state replay does not discover all hidden state. Finite interpretation, Torch transactions and JAX admission tests have different domains; unsupported state and rejection remain outcomes. Coop/Enzyme adapters are not implemented. All six application seeds retain functional execution and a legal deferred-enqueue alternative, which changes forward-visible state while matching the permitted step boundary. Observed output matches and synthetic pressure benefits establish no broad training acceleration.

\paragraph{Why the budget changes the useful policy.}
The sparse-write results describe different choices at different budgets,
not a single universally preferable schedule. At 32 GiB, the joint policy
retains three segments eagerly and replays thirteen with regions. At 44 GiB,
it retains thirteen eagerly and replays three with regions; the correctly
costed eager/whole control instead retains three eagerly and replays
thirteen with whole snapshots. Cheaper historical-state recovery therefore
allows the larger budget to be spent on avoiding more recomputation. The
completed three-workload cohort provides a complementary control: regions
reduce the cost of replay relative to whole snapshots, yet eager retention
is already both faster and smaller there. Improving a replay primitive does
not establish that selecting replay improves the complete execution.
\paragraph{Interpreting application resource and quality observations.}
The dense-queue comparison asks whether that policy-space benefit transfers
to complete training, rather than whether its replay contract is executable.
All seven methods preserve the observed within-seed test-score files, while
whole snapshots consistently use less allocated memory and encoder-time
ordering depends on the seed. The frozen joint schedule retains four
segments eagerly and replays three with regions; whole snapshots and fixed
regions preserve a different activation-retention pattern. Complete-process
peaks cannot be attributed to a single queue copy or program phase because
the counters retain only the largest value over the full observation
interval. Training seeds, rather than test images or allocator resets, define the
independent application units. Paired
seed observations remain descriptive under shared-host contention; the
separate fifty-process sparse-write intervals retain their own frozen
workloads and do not provide uncertainty estimates for the application.

\section{Conclusion}
Valid recomputation preserves value versions, aliased bytes and randomness together. Explicit state costs then change which activation policies fit. Finite and framework checks establish the supported contract; sparse/dense controls and independent-process cohorts delimit its useful regime. Selective regions improve the correctly costed policy space at 44 GiB and expose a speed--memory tradeoff at 32 GiB. All seven methods preserve identical within-seed outputs across six dense-queue seeds. Whole snapshots use less allocated memory in all five observed training seeds; encoder-time ordering reverses at one seed. The dense application thus exposes a resource tradeoff, while the confirmed sparse-write benefit holds within its own workload scope. Together they give a concrete rule for the studied workloads: selective regions pay off when writes are sparse and activations are large, eager retention remains competitive when it fits, and whole snapshots used less allocated memory in the dense-queue application.

The contract and planner open several directions for future work. Replay is restricted to declared state, so automatic discovery of arbitrary aliases and hidden buffers remains outside the contract. Coop and Enzyme adapters are not implemented, and offloading options such as ZeRO-Offload \citep{offload} lie outside the static policy space; neither is ruled out by the reported infeasibility results. Budgets are logical rather than enforced, and the static model does not establish allocator-level optimality. Extending the planner to allocator-aware costs, placement and communication, and to stateful programs whose writes are not declared in advance, would test whether the regime boundary observed here between sparse-write and dense-queue workloads persists in a broader policy space.

\section*{AI Usage Disclosure}
We used ChatGPT (OpenAI) to refine the manuscript's language and formatting, and Claude (Anthropic) to assist with figure preparation. AI assistance was limited to these presentation-related tasks and did not extend to research ideation, literature review, experimental design, research code generation, or the development of technical content. We reviewed and approved all AI-assisted edits and figures and take full responsibility for the final manuscript.

\section*{Ethics Statement}
The artifact retains complete outcomes, unsupported cases, deliberate negative controls, seed identities and source hashes. Framework rejection, semantic divergence, static infeasibility and measured performance use separate categories. The experiments contain no personal data, and computed policy costs are never substituted for hardware observations.

%% file: teaser_figure.tex
\begin{figure}[t]
\centering
\includegraphics[width=\columnwidth]{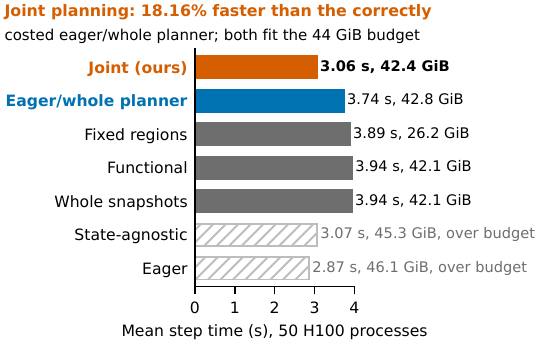}
\caption{Headline result. On the frozen sparse-write H100 workload with a 44 GiB logical budget, joint activation/state planning reduces mean step time by 18.16\% relative to a correctly costed planner restricted to eager retention and whole snapshots (approximate 97.5\% ratio interval $[.8172,.8195]$, 50 independent processes). Hatched bars exceed the logical budget. Labels give mean step time and peak allocation.}
\label{fig:teaser}
\end{figure}

%% file: preservation_engines.tex
\begin{table*}[t]
\caption{Five preservation engines in the original 1,200-program runtime study. They operate on the declared finite floating-point program domain. Integer-write differentiation is rejected by the separate joint-alias admission check; none implements general alias discovery or concurrent external effects.}
\label{tab:engine-contracts}
\centering\small
\begin{tabularx}{\textwidth}{@{}p{.14\textwidth}XXX@{}}
\toprule
Engine & Retained state & Recovery for backward & Supported boundary\\
\midrule
Eager retention & Full pre-operation states & Read the saved prestate & Explicit finite storage and recorded operation descriptors\\
Whole checkpoint & Periodic full-state anchors and recorded random inputs & Replay from the preceding anchor to the required operation & Replay uses the original noise/mask records\\
Read-set checkpoint & Historical values read by each operation & Read the retained operands directly & Read indices are declared by the finite program\\
Region journal & Old and new values of overwritten destinations & Undo writes in reverse order; redo writes to restore the final state & Destinations and write order are explicitly recorded\\
Guarded region & The selected journal or whole-anchor representation & Use whole checkpoints when journal payload exceeds anchor payload; otherwise undo/redo & A byte-count selection rule, rather than measured latency optimization\\
\bottomrule
\end{tabularx}
\end{table*}

%% file: stateful_replay_architecture_v8.tex
\begin{figure*}[t]
\centering
\includegraphics[width=\textwidth]{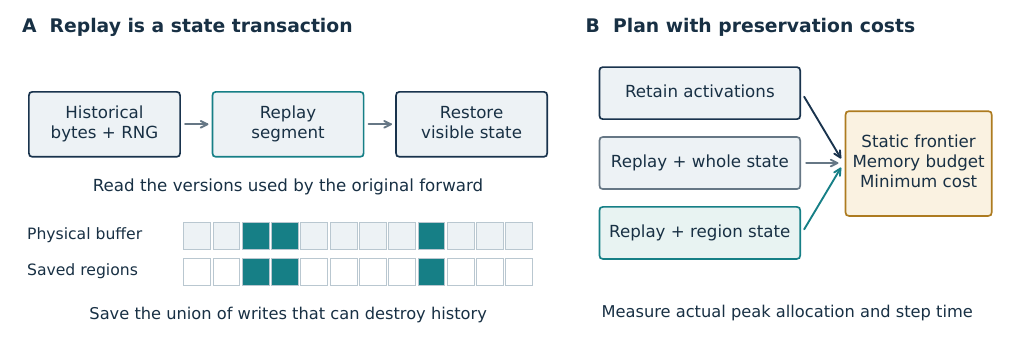}
\caption{Stateful recomputation couples replay correctness to resource planning. (A) Replay temporarily restores historical declared bytes and random-generator state, executes the segment, then reinstates the externally visible state. Selective preservation covers the union of writes that could destroy required history, including later writes. The colored buffer cells are schematic. (B) The planner charges activation retention, temporary workspace and preservation/recovery cost for eager, whole-state and region-state options before selecting a policy under a logical budget. Measured allocation and time evaluate the selected policy.}
\label{fig:stateful-replay-architecture}
\end{figure*}

%% file: framework_torch_results.tex
\section{Replay Semantics in PyTorch and JAX}
Fifty seed keys (420000--420049) generate five paired program families: local strided writes, local overlapping views, hidden-buffer strided writes, hidden-buffer overlapping views, and raw overlapping copy. The same program families run in FP32 and FP64 on H100. The completed cohort contains 250 input programs and 5,900 execution records: 3,500 Torch and 2,400 JAX observations. Methods, precisions and families remain paired within the seed; individual records are not independent replications.

The reference is a functional execution of the same program. Torch 2.8.0+cu126 eager, reentrant checkpointing, nonreentrant checkpointing and transactional nonreentrant replay are tested. The same forward graph is differentiated twice. Native checkpoint RNG preservation is enabled except in named negative controls; nonreentrant early stopping retains its default. The transaction restores declared hidden buffers and random state around replay, then restores externally visible state. It does not discover arbitrary hidden state automatically.

\begin{table}[t]
\caption{Complete framework observation cohort. M: matches recorded reference state and backward observations; R: framework rejection; D: observed divergence. Every outcome remains in its row denominator. These are paired conditions from 50 seed keys, not 5,900 independent trials.}
\label{tab:framework-native}\centering\small
\setlength{\tabcolsep}{3pt}
\begin{tabularx}{\linewidth}{@{}Yrrr@{}}\toprule
Backend / execution & M & R & D\\\midrule
Torch functional reference &500&0&0\\
Torch native eager &400&100&0\\
Torch reentrant &200&100&200\\
Torch nonreentrant &200&100&200\\
Torch reentrant, RNG unpreserved &0&100&400\\
Torch nonreentrant, RNG unpreserved &0&100&400\\
Torch transactional nonreentrant &400&100&0\\\midrule
JAX functional reference &500&0&0\\
JAX functional checkpoint &500&0&0\\
JAX native in-place eager &0&500&0\\
JAX native in-place checkpoint &0&500&0\\
JAX impure-holder eager &200&0&0\\
JAX impure-holder checkpoint &0&200&0\\
\bottomrule\end{tabularx}
\end{table}

Table~\ref{tab:framework-native} distinguishes preserved execution, explicit rejection and an unnoticed mismatch. Torch's two native checkpoint variants match all 200 local-view conditions but diverge on all 200 hidden-buffer conditions; the declared-state transaction matches both families. The 100 raw-copy conditions are rejected even by eager execution. These stateful functions are outside an ordinary pure-function checkpoint contract, so the result does not imply a framework violates its documented assumptions. Disabling RNG preservation produces divergence on all 400 non-copy conditions.

JAX functional checkpointing matches all 500 functional conditions. Native in-place execution is rejected, and checkpointing the 200 impure-holder conditions is also rejected. Those rejections are evidence of a supported-domain boundary, not silently incorrect gradients. The cross-framework result is therefore a distinction in admission and state handling, rather than a claim that both frameworks exhibit the same failure.

Full observations, invocation counts, exceptions and source hashes are retained. A local checker artifact includes pinned requirements, commands, the 5,900-row cohort and an independently derived summary. Archive integrity and its command-line entry point were checked; a fresh dependency installation and a public repository release are separate tasks. This semantic study does not contribute model-training or throughput measurements.

%% file: evidence_results.tex
\section{Independent Engines and Joint Alias Stress Tests}
The semantic study executes 1,200 programs spanning queue, prototype, and recurrent workloads. Five supported preservation engines include eager retention, whole-buffer checkpoints, read-set checkpoints, region journals, and guarded regions. Stored operation traces permit an independent gradient reconstruction. Across the supported engines, maximum gradient error is $1.56\times10^{-15}$. A separate complex-step check covers 96 programs and 4,544 coordinates, with maximum discrepancy $1.00\times10^{-15}$.

\begin{table}[t]\caption{Semantic validation layers. Counts refer to executed programs or method--program outcomes, not hardware benchmark repetitions.}\label{tab:replay-contract}\centering\small
\begin{tabularx}{\linewidth}{@{}Yr@{}}\toprule
Validation quantity & Result\\\midrule
Independent semantic programs &1,200\\
Complex-step coordinates &4,544\\
Maximum complex-step error &$9.99\times10^{-16}$\\
Joint-alias programs &300\\
Supported joint recoveries &1,500/1,500\\
Maximum joint gradient error &$4.44\times10^{-16}$\\
Explicit unsupported outcomes &300\\\bottomrule
\end{tabularx}\end{table}

The joint-alias study uses 50 seeds, three workload patterns, and dimensions 32/64. Each of its 300 inputs combines floating storage with integer reads, strided aliases, partial overlaps, mutation, and replay. The 4,800 stored operations expose byte offsets and before/after values. All 1,500 supported method--program recoveries agree in restored bytes, random state, alias storage, and repeated backward results. The additional 300 integer-write/float-gradient outcomes are marked unsupported rather than assigned fabricated gradients.

These experiments broaden the earlier isolated hazards into a joint contract test. They also distinguish semantic execution from the accompanying cycle and capacity accounting: those resource quantities are calculated under specified models, while the gradient, byte, alias, and replay checks above are direct execution outcomes.

%% file: cost_calibration_results.tex
\section{When Selective Preservation Saves Resources}
We complete a mechanism-driven H100 calibration over the prespecified factorial grid: buffers of 64 or 1,024 MiB, write fractions 0.001 and 1, 1 or 64 live versions,4,096-byte granularity, and feature-queue, prototype-memory and unrolled-recurrence workloads. Each shape retains eager, whole-snapshot, region, functional, generic-planner and joint-planner execution. Matching shape-specific H100 profiles select policies under a fixed20 GiB profile-model budget, chosen before comparative outcomes. That budget is not an allocator cap.

All24 shapes execute all six supported methods with successful FP32/FP64 semantic checks. Each calibration process uses 3 warmups and 10 measured steps. The record includes actual allocator peaks, synchronized whole-step time, region indices, planning/initialization costs and both repeated-backward checks. Coop and Enzyme adapters are explicitly unsupported rather than relabeled as an executed baseline. These are calibration results, not the registered50-process experiment.

Large buffers with sparse writes expose the benefit of preserving changed regions. At1,024 MiB, write fraction0.001 and 64 live versions, region preservation uses 5.05\%,15.04\%, and 6.48\% of whole-snapshot peak memory for the three workloads. Its whole-step times are38.99\%,60.79\%, and 48.82\% of the corresponding snapshot times. Thus the queue case reduces snapshot peak by 94.95\% and step time by 61.01\% under the same supported replay contract.

The full grid (Table~\ref{tab:cost-calibration-grid}, Appendix~\ref{app:grid}) identifies the operating boundary. At write fraction1, region and whole-snapshot peak memory are equal, and region bookkeeping adds time. Eager execution is faster and uses less peak memory than region preservation in every calibrated shape; both profile planners select eager throughout the fixed20 GiB budget. Consequently these results support changed-region preservation as an alternative to whole snapshots when recomputation is required, rather than a universal benefit from checkpointing or evidence that joint planning improves an already feasible eager execution. Table~\ref{tab:cost-calibration-grid} retains all sparse/dense and single/multiversion controls.

%% file: activation_pressure_results_v6.tex
\section{Activation Pressure Changes the Feasible Policies}
The original grid leaves eager retention feasible and preferable. A separate four-cell development calibration increases activation pressure while retaining sparse and dense controls. All cells use a 1 GiB feature queue, sixteen live segments and 4,096-byte regions. Small activations use width/batch 256/256; large activations use 4,096/65,536. Write fractions are $.001$ and $1$. Matching H100 profiles choose policies under a fixed 32 GiB logical budget. Each executable method receives three warmups, ten measured steps and FP32/FP64 semantic checks. This is an activation-pressure stress study, not a natural application batch-size recommendation.

\begin{table*}[t]
\caption{All four activation-pressure calibration cells. Entries are measured peak allocated GiB / mean synchronized step milliseconds. Generic denotes the state-agnostic planner; joint includes state costs and selective regions. ``Infeasible'' is the exact static-model planning outcome, not an observed incorrect gradient or device OOM. Each cell is one calibration process.}
\label{tab:activation-pressure}\centering\small
\setlength{\tabcolsep}{3pt}
\begin{tabularx}{\textwidth}{@{}Yrrrrrr@{}}\toprule
Activation / writes & Eager & Whole & Regions & Functional & Generic & Joint\\\midrule
Small / sparse &1.22 / 8.73&18.21 / 65.39&1.35 / 25.38&18.21 / 69.75&1.22 / 8.69&1.22 / 8.70\\
Small / dense &2.38 / 38.37&19.38 / 126.29&19.38 / 140.10&19.38 / 130.49&2.38 / 38.40&2.38 / 38.43\\
Large / sparse &46.06 / 2882&42.06 / 3937&26.20 / 3908&42.06 / 3960&42.81 / 3750&29.94 / 3698\\
Large / dense &46.08 / 2921&42.33 / 4008&42.33 / 4049&42.33 / 4043&43.08 / 3828&Infeasible\\
\bottomrule\end{tabularx}
\end{table*}

In the large sparse cell, joint planning retains three segments eagerly and replays thirteen with region preservation. Its measured peak is 29.936 GiB, 30.08\% below the state-agnostic planner's 42.813 GiB; its 3,698 ms mean step is 5.37\% below fixed region preservation. Eager remains fastest but uses 46.063 GiB. These results identify a memory--time tradeoff under activation pressure. They do not establish a statistically confirmed 5.37\% speed gain from ten nested steps.

The small cells select eager in both planners. Dense large-state writes remove the cheap region option, and the joint static model has no policy under 32 GiB. Every executed policy passes the stated semantic checks. The failed planning admission is retained separately from semantic execution. The H100 NVL has substantially more physical memory than the logical budget; no hardware capacity cap was enforced and no actual OOM avoidance is claimed.

\subsection{A correctly costed whole-snapshot control}
The state-agnostic planner is a diagnostic control for omitted preservation costs, not the strongest generic checkpoint baseline. We therefore evaluate the same static model with state costs included but with its choices restricted to eager retention and whole snapshots. Exhaustive enumeration of all $2^{16}$ policies on each fixed H100 profile gives the minimum predicted peaks in Table~\ref{tab:state-aware-feasibility}. This is a read-only calculation on measured profile cards, with no frontier truncation; it adds no GPU timing observations.

\begin{table}[t]
\caption{Correctly costed eager/whole policies under the same static profile model. These are predicted minima, not measured device peaks. The sparse large cell admits a joint region policy at 32 GiB but no eager/whole combination.}
\label{tab:state-aware-feasibility}\centering\small
\begin{tabular}{@{}lrr@{}}\toprule
Cell & Minimum predicted GiB & Fits 32 GiB?\\\midrule
Small sparse &1.2211&Yes\\
Small dense &2.3887&Yes\\
Large sparse &40.0631&No\\
Large dense &40.0786&No\\
\bottomrule\end{tabular}
\end{table}

This establishes the additional feasible policies supplied by region preservation within the stated model. It does not exclude other schedules, offloading or a semantically equivalent application rewrite. At a separately frozen 44 GiB budget, both the correctly costed eager/whole planner and joint planner select feasible policies. Section~\ref{sec:fair44} reports their separate hardware calibration.

%% file: fair44_results_v1.tex
\section{A Fair Measured Control at 44 GiB}
\label{sec:fair44}
The 32 GiB comparison establishes a difference in the feasible policy sets. To compare execution cost when both planners have an admissible policy, we freeze an additional 44 GiB logical budget on the same large/sparse H100 profile. The state-aware eager/whole planner uses exactly the same retained-state, temporary-state and preservation-time costs as the joint planner; only its option set excludes regions. Its policy is checked against exhaustive enumeration of all $2^{16}$ choices. Both dynamic-programming frontiers are untruncated.

The correctly costed control retains three segments eagerly and replays thirteen with whole snapshots. The joint policy retains thirteen eagerly and replays three with region preservation. This is the reverse of the earlier 32 GiB joint allocation, as the larger budget permits more activation retention. The generic planner remains as an omitted-state-cost diagnostic. All seven methods execute in one H100 process with three warmups and twenty measured steps per method; every method passes the FP32/FP64 semantic checks. The 140 timing rows and their source/profile hashes are retained.

\begin{table}[t]
\caption{Separate 44 GiB fair-control calibration. Peak is measured allocated GiB; time is mean synchronized milliseconds per step. This is one process, not fifty independent launches. Fits refers to the logical budget, not an enforced physical limit.}
\label{tab:fair44}\centering\small
\setlength{\tabcolsep}{3pt}
\begin{tabular}{@{}lrrc@{}}\toprule
Method & Peak GiB & Step ms & Fits?\\\midrule
Eager &46.063&2865.83&No\\
Whole snapshots &42.063&3919.70&Yes\\
Fixed regions &26.200&3867.19&Yes\\
Functional &42.063&3932.56&Yes\\
State-agnostic planner &45.313&3068.25&No\\
State-aware eager/whole &42.813&3735.42&Yes\\
Joint planner &42.352&3053.21&Yes\\
\bottomrule\end{tabular}
\end{table}

Relative to the correctly costed eager/whole planner, joint planning reduces measured mean step time by 18.26\% and peak allocated memory by 1.08\%. Both fit 44 GiB. Eager remains faster but exceeds the budget, and fixed regions use less memory but take longer. The result isolates the value of choosing region preservation jointly with activation retention in this calibration; it does not rely on an incorrectly costed baseline to establish the timing comparison.

Static predictions remain distinct from measurements. The state-aware control predicts 43.813 GiB and 3,748.53 ms, while joint planning predicts 43.352 GiB and 3,085.27 ms. Their measured peaks are about one GiB lower, illustrating the approximation in the profile model. The H100 NVL has substantially more physical memory than 44 GiB, so the comparison makes no claim of an observed device OOM. Section~\ref{sec:confirmation} reports separate independent-process confirmation; the dense-queue application tests transfer under a different state-access pattern.

%% file: confirmation_results_v1.tex
\section{Independent-Process Confirmation}
\label{sec:confirmation}
Two completed cohorts freeze the large-activation, sparse-write profile before confirmation. Each launches fifty fresh processes on H100 NVL with Torch 2.8.0+cu126 and CUDA 12.6. The workload has a 1 GiB feature queue, write fraction $.001$, sixteen live segments, 4,096-entry region granularity, width 4,096 and batch 65,536. The shared profile digest is fixed across both cohorts. Each process executes all methods for its cohort with three warmups and twenty synchronized measured steps per method. The 44 GiB cohort retains seven methods and 7,000 raw timing rows; the 32 GiB cohort retains six methods and 6,000 rows. Every method passes the FP32/FP64 semantic checks in every process. Terminal status, source identity and the archived raw timings were verified before analysis.

\begin{table*}[t]
\caption{Complete confirmation outcomes. Time is the arithmetic mean of fifty process means in milliseconds; peak is allocated GiB and is identical across the fifty processes for each method. Fits counts observed peaks within the stated logical budget. The two cohorts are distinct and neither enforces an allocator cap. The eager/whole planner is not executed at 32 GiB, where its static policy set is infeasible.}
\label{tab:confirmation-all}
\centering\small
\setlength{\tabcolsep}{5pt}
\begin{tabular}{@{}lrrrrrr@{}}
\toprule
&\multicolumn{3}{c}{44 GiB cohort}&\multicolumn{3}{c}{32 GiB cohort}\\
Method&Step ms&Peak GiB&Fits/50&Step ms&Peak GiB&Fits/50\\
\midrule
Eager&2869.65&46.063&0&2878.70&46.063&0\\
Whole snapshots&3940.38&42.063&50&3945.99&42.063&0\\
Fixed regions&3889.79&26.200&50&3898.82&26.200&50\\
Functional&3942.81&42.063&50&3952.28&42.063&0\\
State-agnostic planner&3070.72&45.313&0&3744.73&42.813&0\\
State-aware eager/whole&3740.91&42.813&50&---&---&---\\
Joint planner&3061.39&42.352&50&3704.54&29.936&50\\
\bottomrule
\end{tabular}
\end{table*}

Inference pairs methods within process. The point estimate is the geometric mean of fifty process ratios; 20,000 paired-process bootstrap draws supply approximate 97.5\% pointwise intervals for the two prespecified primary comparisons in each cohort. Nested steps and the earlier development calibration do not increase the independent sample size. At 44 GiB, the joint/eager--whole time ratio is $.818351$ with interval $[.817193,.819465]$, a reduction of 18.16\%. Their peak ratio is $.989239$, or a 1.08\% reduction. Both methods fit in all fifty processes; eager is faster but exceeds 44 GiB throughout.

At 32 GiB, joint planning and fixed regions fit in every process. Their time ratio is $.950163$ with interval $[.949090,.951220]$, a 4.98\% reduction, while joint planning uses 29.936 GiB versus 26.200 GiB. The joint/state-agnostic peak ratio is $.699226$; that comparator exceeds 32 GiB in every process. It diagnoses an omitted-cost policy rather than providing a feasible timing competitor. Identical observed allocation peaks give point-valued bootstrap intervals for the memory ratios; this does not imply zero uncertainty under other shapes, allocators or hardware. No infeasible policy is assigned an infinite measured runtime, and no logical-budget exceedance is described as a device OOM.

%% file: original150_results_v2.tex
\paragraph{Three-workload confirmation.}
The frozen three-cell cohort comprises 150 independent launches:
feature queue, prototype memory, and unrolled recurrence, each with a
1024\,MiB state buffer, requested write fraction $0.001$, and 64 live
segments. Each of the six methods has 50 warm-up and 1000 measured steps
per launch. All 900 method--process observations completed, and the FP32
and FP64 semantic checks passed. The 900,000 nested timing rows do not
increase the independent sample size beyond 50 launches per workload.
Table~\ref{tab:original150-primary} reports the six prespecified paired
comparisons, recomputed from the raw rows using 20,000 process-bootstrap
draws and Bonferroni adjustment across the six comparisons.

\begin{table}[t]
\centering\small
\caption{Regions/whole geometric-mean ratios in the completed cohort.
Brackets give the adjusted paired process-bootstrap intervals. Point
intervals reflect identical observed allocator values, not absence of
population uncertainty. Lower is better.}
\label{tab:original150-primary}
\begin{tabular}{lcc}
\toprule
Workload & Step time & Peak allocated memory\\
\midrule
Queue & .3954 [.3905,.4020] & .05055 [.05055,.05055]\\
Prototype & .6085 [.6078,.6091] & .15035 [.15030,.15039]\\
Recurrence & .4895 [.4890,.4900] & .06476 [.06476,.06476]\\
\bottomrule
\end{tabular}
\end{table}

These reductions establish the cost of avoiding complete state copies
when replay is selected; they do not establish that replay is best for
these cells. Eager retention is both faster and smaller than every
uniform replay method here. Both planners select all-eager policies on
these activation shapes, and their measured costs are correspondingly
similar. All six method records remain available in the artifact. The complete joint planner
therefore has no demonstrated advantage over the generic planner in
this cohort. The separate activation-heavy 44\,GiB and 32\,GiB cohorts
remain the evidence for their respective planner comparisons; we do not
pool them with this cohort. These three cells characterize their stated low-activation workloads; the
separate pressure and application cohorts test different resource regimes.

%% file: contrastive_four_seed_results_v2.tex
\section{Dense-Queue Application Evaluation}
The application trains CIFAR-adapted ResNet18 query and momentum encoders
on CIFAR-100 using two-view InfoNCE and a 64 MiB FP32 negative queue.
Strict execution enqueues during forward; replay must preserve those queue
values. Deferred eager enqueues after backward and update. With one
outstanding forward, it preserves the step-boundary computation while
changing forward-visible timing, and remains a legal control.

All seven methods execute 100 encoder and 25 frozen-probe epochs at each
of six seeds, 2701--2706: 42 paired method--seed conditions. Every run uses the fixed 40,000/10,000 fitting/validation
split, four million fitting-image visits and eight million augmented views.
The probe is hash-bound to its final encoder checkpoint; test data select no
epoch. All seven full test-score files are byte-identical within each seed.
Independent CPU recomputation yields accuracies 27.16\%, 23.70\%, 26.62\%,
27.40\%, 23.75\% and 26.44\%, and mean cross-entropies 2.9437586983,
3.1000932514, 2.9622467268, 2.9076578604, 3.1101982098 and 2.9579001328,
respectively, on 10,000 official test images. The independent quality unit is a trained seed ($N=6$), not the 42 method
rows or individual test images; within-seed equality does not establish
population equivalence.

\paragraph{Five observed complete encoder processes per method.}
Seeds 2702--2706 capture allocator peaks before all 31,300 resets and once
at exit. Observation spans initialization, encoder training, validation and
checkpoint work, excluding the probe. Counter-read costs are included in
complete encoder time. Allocated and reserved peaks are distinct from
whole-device memory; uninstrumented seed 2701 supplies quality evidence only.
\input{observed_full_memory_table_v2}

Across all five observed seeds, joint planning uses 1,229.74 MiB allocated
versus 1,389.41 MiB for deferred eager, an 11.49\% reduction. Whole snapshots
instead use 1,004.62 MiB. Joint planning uses 22.41\% more allocated memory
than whole snapshots. Its complete-encoder time ratio has geometric mean
1.008024 and seed range $[0.984483,1.019058]$: joint planning is slower at
four seeds but 1.55\% faster at seed 2706. Against deferred eager, its
complete-process ratio is 1.007065, with range $[0.995213,1.020972]$.
Regions likewise show no full-process memory advantage over whole snapshots.
These observations describe a workload-specific tradeoff, not uniform
application acceleration or a joint-planner Pareto improvement over whole
snapshots. All five paired processes are plotted, without a cross-seed
confidence interval or significance claim.

The policies match seed 2701: the state-agnostic planner chooses five whole
snapshots and two eager segments; joint planning chooses three region snapshots
and four eager segments. The frozen 1,024 MiB selection budget constrains the
static model, not the allocator. No observation identifies which program phase
sets a maximum. Sparse-write confirmation cohorts remain separate.

%% file: observed_full_memory_table_v2.tex
\begin{table*}[t]
\centering\small
\caption{All five observed dense-queue training seeds (2702--2706). Allocated and reserved maxima are identical across these seeds for each method. Encoder time includes the observer cost; the range retains all five paired processes and is not a confidence interval. Probes are separate.}
\label{tab:observed-full-memory-three}
\begin{tabular}{lrrrr}
\toprule Method & Allocated MiB & Reserved MiB & Mean encoder seconds & Min--max seconds\\
\midrule
Eager & 1389.35 & 1924.00 & 1857.47 & 1844.03--1866.87\\
Whole & 1004.62 & 1410.00 & 1841.10 & 1821.24--1864.30\\
Regions & 1004.74 & 1410.00 & 1842.40 & 1834.85--1854.06\\
Functional & 1004.62 & 1410.00 & 1847.51 & 1837.14--1862.65\\
Deferred & 1389.41 & 1924.00 & 1842.81 & 1833.59--1850.69\\
Agnostic & 1005.48 & 1410.00 & 1856.22 & 1841.82--1882.82\\
Joint & 1229.74 & 1602.00 & 1855.90 & 1835.37--1878.18\\
\bottomrule
\end{tabular}
\end{table*}

%% file: stateful_replay_evidence_v8.tex
\begin{figure*}[t]
\centering
\includegraphics[width=\textwidth]{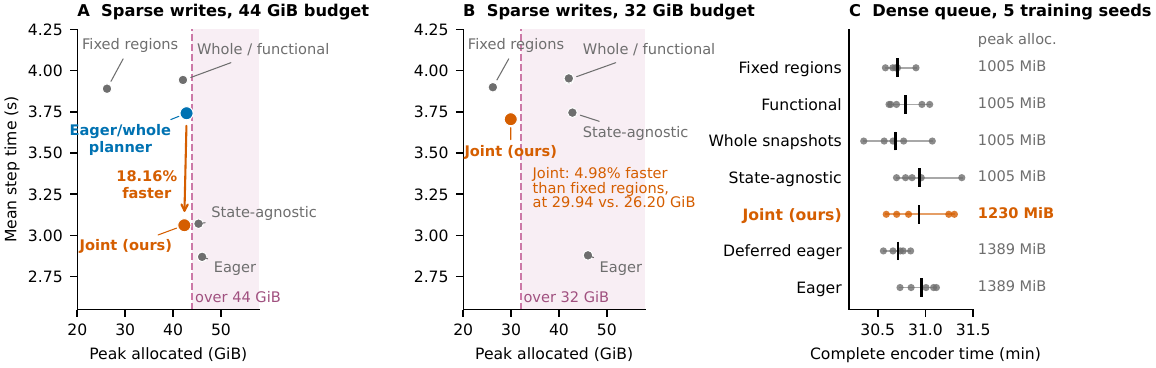}
\caption{The resource benefit depends on the workload and comparator. Joint planning is highlighted in orange and the correctly costed eager/whole control in blue. (A,B) All executed methods in the two frozen large-activation, sparse-write cohorts. Points are means over 50 independent H100 processes; peak allocation is identical across processes for a method. Shading marks allocation above the logical, unenforced budget, not a device out-of-memory event. The correctly costed eager/whole planner is statically infeasible at 32 GiB and has no measured point there. (C) All 35 complete encoder processes: seven methods paired at five observed training seeds (2702--2706), including initialization, 100 encoder epochs, validation and checkpoint work. Probe execution is separate. Each dot is one process; black ticks mark the five-seed mean and labels give the (seed-invariant) peak allocation. Whole snapshots use less allocated memory at all five seeds. Joint planning is faster at seed 2706 and slower at the other four. Seed 2701 supplies quality/execution evidence but no comparable full-process memory replicate. No application confidence interval or significance is inferred.}
\label{fig:stateful-replay-evidence}
\end{figure*}

%% file: cost_calibration_grid.tex
\begin{table*}[t]
\caption{Complete prespecified H100 cost calibration: all 24 shapes and six supported methods execute successfully. Ratios are region preservation divided by whole snapshots or eager execution. Time is the mean of 10 nested measured steps after 3 warmups; peak is the actual maximum allocated memory. Each cell is one process, not 50 independent launches. The full artifact also retains functional and both planner outcomes.}
\label{tab:cost-calibration-grid}\centering\small
\begin{tabular*}{\textwidth}{@{\extracolsep{\fill}}lrrrrrrr@{}}\toprule
Workload & MiB & Write (\%) & Live & Peak/whole & Time/whole & Peak/eager & Time/eager\\\midrule
Queue & 64 & 0.1 & 1 & 0.669 & 1.134 & 1.002 & 2.263 \\
Queue & 64 & 0.1 & 64 & 0.099 & 1.189 & 1.353 & 3.292 \\
Queue & 64 & 100 & 1 & 1.000 & 1.405 & 1.497 & 2.752 \\
Queue & 64 & 100 & 64 & 1.000 & 1.199 & 11.660 & 3.273 \\
Queue & 1024 & 0.1 & 1 & 0.376 & 0.443 & 1.002 & 2.227 \\
Queue & 1024 & 0.1 & 64 & 0.051 & 0.390 & 2.612 & 2.965 \\
Queue & 1024 & 100 & 1 & 1.000 & 1.105 & 1.937 & 3.562 \\
Queue & 1024 & 100 & 64 & 1.000 & 1.111 & 21.407 & 3.632 \\
\midrule
Prototype & 64 & 0.1 & 1 & 0.717 & 1.050 & 1.003 & 2.197 \\
Prototype & 64 & 0.1 & 64 & 0.193 & 1.190 & 2.232 & 3.231 \\
Prototype & 64 & 100 & 1 & 1.000 & 1.132 & 1.398 & 1.552 \\
Prototype & 64 & 100 & 64 & 1.000 & 1.168 & 10.141 & 2.851 \\
Prototype & 1024 & 0.1 & 1 & 0.526 & 0.583 & 1.004 & 2.197 \\
Prototype & 1024 & 0.1 & 64 & 0.150 & 0.608 & 4.349 & 2.423 \\
Prototype & 1024 & 100 & 1 & 1.000 & 1.061 & 1.489 & 2.691 \\
Prototype & 1024 & 100 & 64 & 1.000 & 1.071 & 13.649 & 2.702 \\
\midrule
Recurrence & 64 & 0.1 & 1 & 0.716 & 1.147 & 1.002 & 2.282 \\
Recurrence & 64 & 0.1 & 64 & 0.111 & 1.190 & 1.295 & 3.267 \\
Recurrence & 64 & 100 & 1 & 1.000 & 1.144 & 1.398 & 2.326 \\
Recurrence & 64 & 100 & 64 & 1.000 & 1.205 & 10.152 & 3.269 \\
Recurrence & 1024 & 0.1 & 1 & 0.524 & 0.495 & 1.001 & 2.183 \\
Recurrence & 1024 & 0.1 & 64 & 0.065 & 0.488 & 1.908 & 2.285 \\
Recurrence & 1024 & 100 & 1 & 1.000 & 1.093 & 1.903 & 3.279 \\
Recurrence & 1024 & 100 & 64 & 1.000 & 1.105 & 20.899 & 3.311 \\
\bottomrule\end{tabular*}
\end{table*}